\documentclass[conference]{IEEEtran}

\IEEEoverridecommandlockouts
\usepackage{cite}
\usepackage{amsmath,amssymb,amsfonts}
\usepackage{algorithmic}
\usepackage{graphicx}
\usepackage{textcomp}
\usepackage{xcolor}
\usepackage{lipsum}
\usepackage{booktabs}
\usepackage{graphicx}
\usepackage{parskip}
\usepackage{tabu}
\usepackage{url}
\usepackage{wrapfig}
\usepackage{multirow}
\usepackage{booktabs}
\usepackage{titlesec}
\usepackage{hyperref}

\def\BibTeX{{\rm B\kern-.05em{\sc i\kern-.025em b}\kern-.08em
    T\kern-.1667em\lower.7ex\hbox{E}\kern-.125emX}}

\begin{document}

\title{Advancements and Challenges in Bangla Question Answering Models: A Comprehensive Review}

\author{\IEEEauthorblockN{Md Iftekhar Islam Tashik, Abdullah Khondoker, Enam Ahmed Taufik, Antara Firoz Parsa, S M Ishtiak Mahmud}
\IEEEauthorblockA{Department of Computer Science and Engineering (CSE)\\ School of Data and Sciences (SDS)\\ 
Brac University\\Kha 224 Merul Badda, Dhaka - 1212, Bangladesh}  
\IEEEauthorblockA{\{iftekhar.islam.tashik, abdullah.khondoker, enam.ahmed.taufik, antara.firuz.parsa, s.m.ishtiak.mahmud\}@g.bracu.ac.bd}}

\maketitle

\begin{abstract}
The domain of Natural Language Processing (NLP) has experienced notable progress in the evolution of Bangla Question Answering (QA) systems. This paper presents a comprehensive review of seven research articles that contribute to the progress in this domain. These research studies explore different aspects of creating question-answering systems for the Bangla language. They cover areas like collecting data, preparing it for analysis, designing models, conducting experiments, and interpreting results. The papers introduce innovative methods like using LSTM-based models with attention mechanisms, context-based QA systems, and deep learning techniques based on prior knowledge. However, despite the progress made, several challenges remain, including the lack of well-annotated data, the absence of high-quality reading comprehension datasets, and difficulties in understanding the meaning of words in context. Bangla QA models' precision and applicability are constrained by these challenges. This review emphasizes the significance of these research contributions by highlighting the developments achieved in creating Bangla QA systems as well as the ongoing effort required to get past roadblocks and improve the performance of these systems for actual language comprehension tasks.

\end{abstract}


\section{Introduction}
Natural Language Processing (NLP) has made great progress in recent years, notably in the creation of question-answering (QA) systems. These systems are critical in extracting useful information from text and giving correct responses to user inquiries. However, there are special obstacles and potential in this field for languages with limited linguistic resources, such as Bangla (Bengali). In this introduction, we will look at the present state of Bangla QA models, as well as the possibilities for future enhancements.
\subsection{Progress in Bangla QA Models:}
During the last decade, there was a substantial rise in the development of NLP models for the Bangla language. Researchers have worked hard to create effective quality assurance systems that can read and interpret Bangla, allowing for more natural interactions with digital platforms. As a result, several methods, approaches, and structures have emerged to deal with the grammatical and contextual difficulties of the Bangla language.

\subsection{Challenges and Issues:}
Considering these advancements, there are still substantial obstacles to developing trustworthy Bangla QA models. One of the most serious issues is a dearth of high-quality labeled datasets for training and evaluating these algorithms. The lack of a large and diversified QA dataset in Bangla has hampered the application of cutting-edge machine-learning approaches. The complicated morphology of the Bangla language, with multiple inflections and syntactic modifications, adds to the challenge of constructing exact systems for interpreting and retrieving textual information. The model's capacity to answer inquiries on a variety of topics is further limited by a lack of domain-specific resources.

\subsection{Language and Cultural Context:}
Building QA models for Bangla requires a deep understanding of the language's linguistic features and cultural context. The inherent structure of Bangla sentences, the presence of honorifics, and the use of context-dependent expressions all contribute to the complexity of modeling. Translating these intricacies into effective algorithms requires a fusion of linguistics and machine learning techniques.

\subsection{Transfer Learning and Multilingual Models:}
Leveraging transfer learning and pre-trained multilingual models has shown promise in mitigating the challenges of low-resource languages like Bangla. By adapting models trained in other languages to Bangla QA tasks, researchers have managed to overcome data scarcity to some extent. Multilingual embeddings and transfer learning techniques facilitate the transfer of knowledge gained from well-resourced languages to the Bangla language domain.

\subsection{Future Directions:}
As research in Bangla QA models progresses, it opens up new avenues for addressing the existing challenges. Developing larger and more diverse datasets, fine-tuning transfer learning models, and exploring domain adaptation techniques are areas of active research. Furthermore, advances in deep learning architectures, attention mechanisms, and contextual embeddings are likely to enhance the performance of Bangla QA models.

\section{Literature review}

Nabi et al.\cite{Nabi} presents a comparison between Bangla and English-based question-answer systems in the domains of International General Knowledge (GK), Bangladesh GK, and Science \& Technology. They propose a Sequence to Sequence LSTM-based approach enhanced with attention mechanisms for improved accuracy. The research focuses on preprocessing data, model architecture, training, and evaluation. In the initial stage, data preprocessing involves organizing datasets with distinct columns for questions, context, and answers. Context creation uses regular expressions, and Bangla questions are paired with answers after removing stopwords. Bangla datasets are also translated into English, and the data is loaded into Colab. The encoder layer tokenizes and pads the data using the pad sequence technique to ensure uniformity in sequence length for effective analysis. The LSTM model, a vital component rooted in the RNN architecture, handles memory limitations and gradient challenges. Its mechanism of using "doors" to manage data retention or disregarding is crucial for NLP and time-series tasks. The LSTM's role within the seq2seq framework is explained, with LSTM blocks present in encoders and decoders. The integration of attention mechanisms empowers decoders to focus on specific input segments. Self-attention layers aid encoder sites in understanding previous layers, enhancing contextual comprehension. Post encoding, softmax normalization of encoded vectors generates a time-dependent outcome fed into the decoder. The model's taxonomy categorizes elements like task delineations, evaluation methodologies, corpus considerations, and question-answer types into five sections. This comprehensive approach contributes to a sophisticated model tailored for advanced question-answering systems. The experimental phase involves training the model over 35 epochs, tracking loss and accuracy. Loss decreases from 2.0494 to 0.1397, and accuracy increases from 0.7829 to 0.9948. The inverse relationship between loss and accuracy is evident as epochs progress. Validation loss (Val\_loss) and accuracy (Val\_accuracy) are also tracked across epochs. Val\_loss starts below training loss but surpasses it by epoch 35, affecting Val\_accuracy, which remains slightly lower than training accuracy. The English model outperforms the Bangla model, achieving higher prediction accuracy (90.16\% vs. 87.36\%). Graphical representations indicate accurate response predictions, although slight discrepancies in real-life user queries could impact performance. To sum up, Nusrat Nabi et al. conducted a comparative study of Bangla and English question-answer systems, proposing a Sequence to Sequence LSTM-based model with attention mechanisms. The research covers data preprocessing, model architecture, and training stages. The LSTM's role in the seq2seq framework and the importance of attention mechanisms are highlighted. Experimental results show higher accuracy for the English model, emphasizing its effectiveness. The study provides insights into building advanced question-answering systems with improved accuracy and performance.

Banerjee et al.\cite{Banerjee} emphasize the building of a factoid question-answering system for the Bengali language. As the paper presents the first initial attempt to develop a factoid question answering, it also explains the challenges while developing the system. This paper also proposes the extraction and the ranking of relevant sentences. The authors present an automatic system and it can answer natural language questions in a human-like approach. In general, Factoid questions, list questions, definition questions, complex questions and speculative questions are five categories of questions. This system answers factoid questions for three types of Factors. 
The authors also highlight the challenges to developing a question answering for low-resource language. Significant issues that they address include the abundance of interrogatives present in  Bengali, the variety of positions of these interrogatives and the lack of language processing tools.
The authors address the system as BFQA (Factoid QA system for Bengali language). It has three components of pipeline architecture. The analysis of  question, sentence extraction and answer extraction are the components. Question type (QType) identification, expected answer type (EAT) identification, named entity identification, question topical target (QTT) identification and keyword identification are the five question analysis step processors. Sentences are extracted and  based on the answer score value, they are ranked. Using the EAT module, extracted answers are validated. The authors create their own corpus for experimentation, as there is no existing Bengali QA corpus available.From Wikipedia, fourteen documents were gathered in the field of  geography and agriculture and it also includes twenty luggage professionals.  184 factoid questions  annotate three levels of  question answering :  Question Class (kappa - 0.91), Expected Answer Type (kappa - 0.85) and Question Topical Target (kappa - 0.89). To evaluate the QA system, Mean Reciprocal Rank (MRR) metric is used. With 10 documents and 139 questions, in geography the Mean Reciprocal Rank (MRR) value is .34 and in agriculture with 4 documents and 45 questions, MRR value is 0.31 . Overall the MRR is 0.32 for the system. This paper discusses the first effort of developing a  factoid question answering system for Bengali language and designing an architecture for low-resource languages particularly for Indian languages. But the testing accuracy of the system is below that of European languages. Because of the  low accuracies of the shallow parser and the NER system, can be the reason for the poor performance of the system as the shallow parser and NER system have a big impact on factoid QA performance.

Bhuiyan et al.\cite{Bhuiyan} constructs an automatic context based Question \& Answering system and discuss the significance of question answering systems in Bangla. Using sequence to sequence architecture,  an attention mechanism is applied following an encoder layer a bi-directional LSTM and decoder.  Data collection, finding the right vocabulary for word mapping were some of their challenges while developing the system. Answering the questions is the main function of the model and it reduces the training loss to 0.003. The authors use a close domain approach in this research.Into two layers LSTMs, input contexts, questions are entered. By following  Bahdanau’s attention mechanism \cite{Attention_Mechanism}, the decoder generates the answer. For data collection, the authors gathered General Knowledge-based question-answers from various websites and collected 1000 question-answers with context. There are three columns in the dataset: context, question and answer. The authors use the LSTM model  for language modeling and handling text data. For machine translation, another type of model is used.  Neural machine translation is a kind of artificial network and it transforms a source language into another. Most machine translations are used with the encoder-decoder where encoder has taken the series of input and decoder generates the output.
The authors develop RNN encoder-decoder. They also used it for machine translation problems. But for vanishing and long sequences RNN doesn’t work well. So the authors apply a Bidirectional LSTM based encoder decoder architecture. Machine translation, text summarization, text generation etc are the artificial intelligence areas where Attention based model have been applied. Contexts, questions are both used as inputs  in the encoding part and entered into the embedding layer. Into BiLSTM both embedding vectors are passed. Another BiLSTM is used for learning from data in the decoding step.After generating the encoded vectors and normalizing them with softmax, using  another BiLSTM the decoder can predict the outcome. After training the training loss of the model reduces to 0.003.

Islam et al.\cite{Islam} is presented for the automatic question-answering system, aiming for proving answers using keyword, lexical and semantic features of a question. Their system provides specific answers for time-related inquiries, while for other questions, it retrieves pertinent information from multiple sources.  The authors present an original Bangla language QA system in NLP, capable of identifying question types and providing specific answers for time and quantity-related questions while retrieving relevant answers for other questions from single or multiple documents. The study addresses QA challenges, outlines system implementation, evaluates its performance, and concludes with findings. During this research, the authors had to face many challenges. Such as: extracting answers from unstructured content might be a difficult task as it might result in a single answer in the source text. Again comparing question answers via lexical, syntactic or semantic relationship is also challenging. The more redundant answer impacts retrieval of answer, otherwise, NLP must address this issue. Lastly, due to the lack of ‘wh’ words in the question, identifying keywords can be difficult. The proposed system in this paper involves preprocessing steps such as removing stop words and stemming from both questions and text documents to enhance matching accuracy. It emphasizes intelligent handling of parts of speech and suffixes, aiming to improve information retrieval and approximate matching performance.For generating keywords or headwords from questions, the author approaches with a statistical method involving an intermediate distance vector and its mean value. Using N-grams (unigram/bigram/trigram), keywords are formed to give approximate matching and help with retrieving relevant information from the source. For Bangla questions, wh-words play a vital role, while ‘|’ says it is a statement. The proposed system employs measurement units as a semantic feature, retrieving answers based on the interrogative (wh-type) type and the semantic attribute. With these retrieved answers, this system will use a Textual Entailment Module (TE) and the highest rank will be regarded as the best answer. The evaluation of the proposed system involves precision, recall and F score calculation using a set of 500 questions and selected wiki documents. The average precision (0.35) and recall (0.65) indicate system performance, leading to an F score of 0.45 showcasing the system’s effectiveness in information retrieval and relevance. As the table illustrates the precision and recall calculation process with sample questions and relevant answers. For every question, the system's retrieval of relevant answers, precision and recall is demonstrated. Where the average precision is 0.50 and average recall is 0.69 from the calculations based on the collected data. With this paper, the authors developed a multi-document QA system for Bangla language and successfully delivers relevant answers, identifies keywords and gives specific responses for time-quantity queries. The system’s Precision, recall and F score of the system are 0.35, 0.65 and 0.45 respectively. Future tasks involve comparing this approach with systems in other languages, exploring different question types, and enhancing relevant answer retrieval by considering pronouns.

M. Keya et al.\cite{Keya} uses universal sentence encoder to embed and measure the similarity between texts using the cosine distance of the text and a deep averaging network is used to find the best similarity. Evaluating the similarity of the model, the  Pearson correlation value is 0.41.  The significance of text and sentence matching, question noting and semantic similarity is clearly noticeable for natural language processing. As for Bengali text processing, the query response system of natural language processing faces some major challenges. Although, the query response system has different methods as well as their limitations.From the references mentioned, the text-to-text similarity and semantic analysis is encoded universally. For the best results, a self-collected dataset of Bengali content is employed, featuring general knowledge from various sources. Using the cosine similarity, the model computes similarity between two texts. Mathematically, it quantifies the cosine transformation of the angle formed between two vectors situated in a multi-dimensional space. The applied architectural approach by her involves embedding, feed forward layers and cosine distance measurement to find similar texts. It consists of the following methods to provide answers based on similarity scores. The application of the universal sentence encoder is for Bangla inquiry similarity estimation, with a transformer encoder adaptation to achieve strong semantic sentence matching performance. PCA(Principle Component Analysis) is a process that usually reduces larger datasets while preserving its actual data essence. It mainly tries to find out what is unique in different embeddings while discarding shared features. It employs covariance matrix, eigenvectors, and eigenvalues to craft a feature vector that retains essential dataset components. Again the cosine distance formula is used in high dimensional context. It identifies similarity between each text comparing frequency vectors, with the cosine value derived from Euclidean dot product formula.The study utilizes Tensorflow Hub, employing a diverse algorithm to measure similarity, particularly for Bengali question text analysis. The model evaluation employs the Pearson correlation coefficient and P value to measure the linear interrelation between variables, yielding a PCC value of 0.41 and P value of 7.69- indicating a positive correlation. The user's question is used to calculate context similarity using a cosine distance model, yielding the most relevant context-output pairs along with answers. The research results showcase the top 5 similar context-answer pairs with corresponding scores. The research mainly focuses on text similarity through semantic alignment. Utilizing context-query similarity, provides better results to find similar texts. Future work aims to expand data, refine semantic matching with complex questions, and enhance automated question-answering systems through score measurement.

Uddin et al. \cite{Mohsin} addresses the critical issue of developing a question-answering (QA) system for the Bangla language, which has been relatively unexplored. The main objective of the study is to propose an effective methodology for constructing a QA system that can accurately handle paraphrased questions with the support of a single line of context. To achieve this objective, the authors employ advanced deep learning techniques, specifically utilizing Long Short-Term Memory (LSTM) and Gated Recurrent Unit (GRU) algorithms \cite{gru}, combined with different word representation strategies: Positioning Encoding (PE) and word embedding using Word2Vec. The methodology of the study involves a structured series of steps. It begins with data collection, where two datasets are utilized. One dataset is translated from an existing source, while another dataset is manually created by the authors, containing historical information. The latter dataset's creation becomes necessary due to the absence of preprocessed datasets in the Bangla language. Subsequently, the collected data undergoes a preprocessing phase, which includes the removal of extraneous symbols, objects, and non-Bangla words. One significant contribution of the study lies in the introduction of Positioning Encoding and Word2Vec techniques for word representation within the LSTM and GRU models. Positioning Encoding assigns a unique position to each word, enhancing the model's understanding of word order. On the other hand, Word2Vec provides a sophisticated semantic relationship between words in a vector space, facilitating more accurate processing of natural language. The study's results are presented through a detailed analysis of various model variations. The proposed approach is evaluated using different activation functions and optimizers. For instance, when applying LSTM with Positioning Encoding and Softmax activation, the average accuracy reaches approximately 86\%. However, the use of Word2Vec representation in the LSTM model yields significantly improved accuracy, achieving 100\%. Furthermore, the authors compare LSTM and GRU architectures and find that GRU, especially when combined with Softmax activation and Word2Vec representation, attains an average accuracy of 98.86\%. The study makes a noteworthy contribution by highlighting the superiority of its proposed approach compared to previous methods, particularly when applied to the translated bAbI dataset. By addressing the lack of a QA system in the Bangla language, the authors open up possibilities for advancements in natural language processing for this language. Nevertheless, the study also acknowledges certain limitations, such as the challenge of handling unsupervised facts and the potential for enhancing the model with larger datasets and additional tasks, such as induction and deduction.

Mayeesha et al. \cite{Mayeesha} focuses on addressing the challenge of building a question answering (QA) system for the Bengali language, a low-resource language with limited annotated data. The authors recognize the significance of QA systems, especially for industry purposes such as chatbots for frequently asked questions. However, the lack of Bengali-specific high-quality QA datasets poses a challenge. The main challenge addressed in the paper is the scarcity of high-quality reading comprehension datasets for the Bengali language. This hinders the development of QA systems for Bengali, as training models from scratch requires large datasets. The authors propose leveraging transfer learning, specifically zero-shot transfer learning, to adapt pre-trained models to Bengali QA tasks.  The paper aims to explore the feasibility of using transfer learning, particularly zero-shot transfer learning, to develop a QA system for the Bengali language. The authors focus on using pre-trained multilingual BERT models and variants like DistilBERT and RoBERTa for the task. Additionally, they aim to assess the performance of their models by conducting a survey involving grade three and four students. The authors adopt a transfer learning approach, focusing on pretrained language models such as multilingual BERT, DistilBERT, and RoBERTa. They first conduct zero-shot learning experiments where these models are adapted from English QA tasks to Bengali without fine-tuning on any Bengali-specific data. Next, they perform fine-tuning on a synthetic Bengali training dataset created by translating a subset of the SQuAD 2.0 dataset from English to Bengali. They use fuzzy answer matching to handle translation errors in the synthetic dataset. The paper presents a comprehensive set of experiments comparing the performance of various models in the zero-shot and fine-tuned settings. The results show that fine-tuning models on the translated Bengali dataset improves performance compared to the zero-shot approach. Among the models tested, DistilBERT achieved the best performance on the synthetic dataset, outperforming RoBERTa and the other models. The authors also conducted a survey involving grade three and four students to compare the performance of the models with human participants. The surveyed children achieved an EM and F1 score of around 50.83 and 67.32, respectively, while the best-performing model achieved an EM and F1 score of 52.5 and 69.21 on the same questions.

Kowsher et al.\cite{Kowsher}, introduced the Bengali Intelligence Question Answering System (BIQAS), focusing on mathematics and statistics within the realm of Bengali Natural Language Processing (BNLP). The research process is divided into three main phases: collecting useful documents, preparing data, and establishing relationships between data and user inquiries. In order to successfully preprocess the data, several corpora have been used. Several algorithms such as Cosine Similarity, Jaccard Similarity, and Naive Bayes, have been used to establish correlations between questions and replies. The TF-IDF model \cite{TF_IDF} is used to convert documents and queries into vectors because the Cosine Similarity method acts on vectors. Singular Value Decomposition (SVD) \cite{svd} approaches have been used to increase efficiency by lowering execution times and spatial complexity. BIQAS has made considerable use of data preparation tools. Pronoun handling and avoiding repetition have both been accomplished using anaphora resolution. Cleaning words entails eliminating symbols like punctuation and stop words, which are words that have little impact on papers or sentences. Levenshtein distance has been used to choose the optimal lemma word between lemmatization methods for particular Bengali words. These methods include DBSRA and Trie. Levenshtein distance has been used to categorize unfamiliar nouns, including names of locations and people, as known or unknown. Synonym handling has been incorporated into BIQAS since it is important for Natural Language Understanding (NLU). For example, "(large)" has been considered synonymous with "(big)," facilitating accurate responses to user queries. The use of the mathematical and statistical techniques in BIQAS was part of the experimental phase. A variety of corpora were used, including queries produced from these documents, stop words, Bengali root words, and informational documents concerning Noakhali Science and Technology University (NSTU). Tasks including cosine similarity (93.22\%), Jaccard similarity (82.64\%), and Naive Bayes classification (91.34\%) showed good accuracy in performance measures. Comparing BIQAS to English chatbots like Neural Conversational Machine (NCM) and Cleverbot showed how advanced BIQAS was in the Bengali chatbot field. The goal of this research endeavor was to develop a Bengali information retrieval bot. With a focus on preprocessing, dimensionality reduction, and building information-question linkages, the technique combined machine learning, mathematics, and statistics. Future improvements could broaden the application of BIQAS across several fields, including voice replies and cutting-edge deep learning methods like Recurrent Neural Networks (RNN) with BNLP. This study establishes the framework for sophisticated Bengali question-answering systems and proves the superiority of BIQAS over current English chatbots.

Ekram et al. \cite{ekram} introduces a vital resource for the Bangla language processing community by presenting a thorough dataset development method and an empirical evaluation of question-answering models. Six essential processes are included in the dataset-building process, which yields a 3000 passage collection of the highest caliber from Bangla Wikipedia. In order to construct lexically and syntactically unique questions from these excerpts, crowd-workers were carefully selected for their language skill. The questions themselves came in a variety of sorts, including factual, causal, confirmation, and list questions, which put the models' understanding of various linguistic nuances to the test. With the use of validation measures like Exact Match (EM) and F1 score, another group of crowd-workers was enlisted to supply the solutions. The distribution of question-answer types throughout the train, validation, and test sets is shown by the dataset analysis. Four different models, including BanglaT5 \cite{BanglaT5}, mT5 \cite{mT5}, BanglaBERT \cite{BanglaBERT}, and mBERT \cite{mBERT}, were adjusted on the BanglaRQA train set to evaluate models for the question-answering task. Each model had its own special skills, but the highest-performing model, BanglaT5, stood out with an Exact Match (EM) score of 62.42\% and an F1 score of 78.11\% on the BanglaRQA test set. This thorough research also examined model performance for various question-and-answer formats, highlighting areas where the models performed well or encountered difficulties. Additionally, analyses of other datasets, such as bn squad, showed the BanglaRQA dataset's usefulness and prospective applicability in a variety of question-answering tasks. The report ends by noting its shortcomings, such as restrictions on computational and human resources, which pave the way for additional investigations into the processing of Bangla. The privacy and fair treatment of annotators were carefully protected throughout the writing process, and authors and reviewers were thanked for their significant contributions to the study. Therefore, "BanglaRQA" stands out as an important tool, filled with carefully chosen passages, questions, and answers, and promising to advance the fields of Bangla reading comprehension-based question-answering and natural language understanding.

T. T. Aurpa et al.\cite{Aurpa}, a model was developed for a question-answering system. To facilitate this endeavour, the author meticulously curated a dataset consisting of 3636 reading comprehension instances, serving as the cornerstone for their comprehensive experimentation. Throughout their research, an array of deep neural network architectures was harnessed for the training process on this dataset. This included the deployment of LSTM (Long Short-Term Memory), Bi-LSTM (Bidirectional LSTM) models coupled with attention mechanisms, RNN (Recurrent Neural Network), ELECTRA, and the influential BERT (Bidirectional Encoder Representations from Transformers). Significantly, the adoption of transformer-based models, notably BERT, demonstrated exceptional promise. During rigorous testing, BERT achieved an impressive accuracy rate of 87.78\%, while during the training phase, it achieved a remarkable 99\% accuracy. The highest accuracy was achieved with BERT, reaching 87.78\% accuracy. This result was obtained by configuring the model with a learning rate of 1e-4, a batch size of 32, and a max sequence length of 484. For the ELECTRA model, the best accuracy attained was 86.5\%. This achievement came with a learning rate of 5e-4, a batch size of 32, and a max-len of 512. To evaluate model performance, the study used accuracy and loss metrics. For BERT, the training accuracy was 98.54\%, while testing accuracy peaked at 87.78\% after 40 epochs. These high accuracy levels are significant in the domain of Bangla Reading Comprehension. Training and validation losses consistently decreased over the epochs. In the case of the ELECTRA model, training accuracy reached 93.0\%, and testing accuracy was 82.52\%. These findings underscore the critical role of hyperparameter optimization in achieving the best possible model performance, highlighting the proficiency of the models in the context of Bangla Reading Comprehension. Future efforts should prioritize expanding dataset diversity, exploring algorithmic enhancements like ensembling, and addressing challenges in transitioning the system to embedded applications. Additionally, improving question type handling and multilingual support should be considered, while acknowledging study limitations like dataset size and evaluation metrics.

\section{Findings}

\begin{figure}[htbp] 
    \centering
    \includegraphics[width=80px]{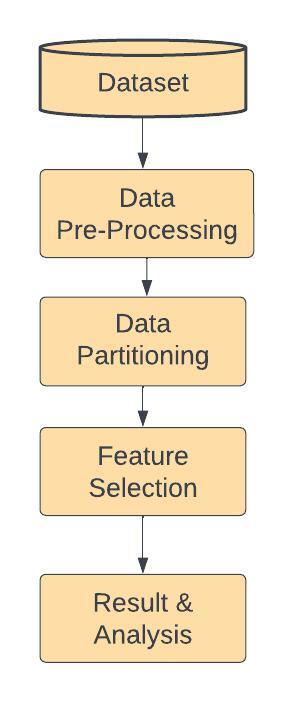} 
    \caption{Text classification approach}
    \label{fig:classification}
\end{figure}

The important conclusions of this investigation were discovered following a thorough analysis and comparison of various scientific works. After carefully reviewing those publications, we discovered that authors in the field often employ a set of four critical methods. Every text classification approach uses these critical techniques: dataset collecting, data pre-processing, data partitioning, and feature selection. The approach shown in Fig.\ref{fig:classification}.

\subsection{Data collection} 
\begin{figure}[htbp] 
    \centering
    \includegraphics[width=0.5\textwidth]{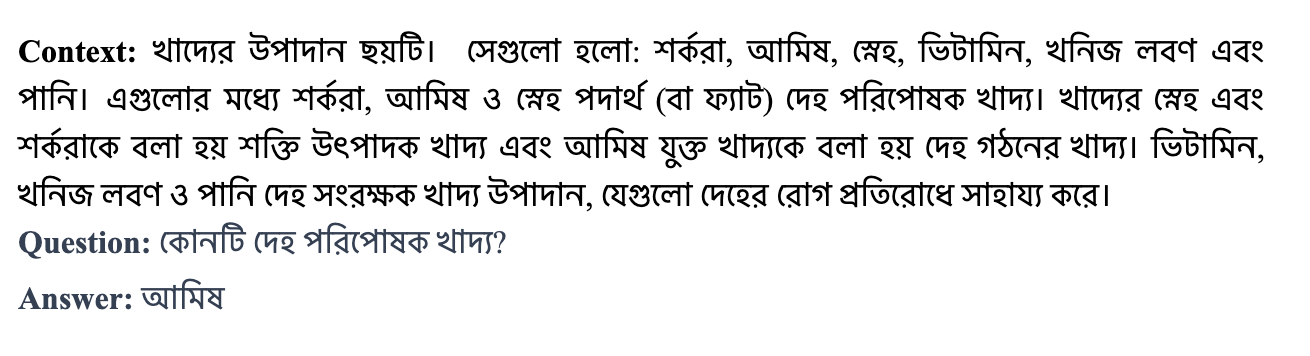} 
    \caption{Sample of Dataset}
    \label{fig:Dataset}
\end{figure}

The data collection methods in the seven papers reviewed vary based on the specific objectives and research approaches of each study. However, there are common themes in data collection processes across these papers. Most of the papers involve the collection of datasets relevant to their respective research topics. This often includes gathering questions, answers, and context information in the Bengali language, This contextual information is essential for training and evaluating question-answering systems as it provides the necessary context for understanding and answering questions accurately. Some papers resort to translating existing datasets from English to Bengali, while others manually create datasets due to the lack of suitable resources. Moreover, some of the datasets also include information about the question types. This categorization of questions into different types is important for analyzing the performance of question-answering systems across various question categories. The sample of dataset is shown in Fig.\ref{fig:Dataset}.

\subsection{Data Preprocessing}

Data preprocessing is a fundamental step in the development of question-answering systems, and the seven papers reviewed employ various techniques tailored to their specific research objectives and dataset characteristics. Fig.\ref{fig:Preprocessing} is an overview of the  data preprocessing methods observed in these papers:

\begin{figure}[htbp] 
    \centering
    \includegraphics[width=270px]{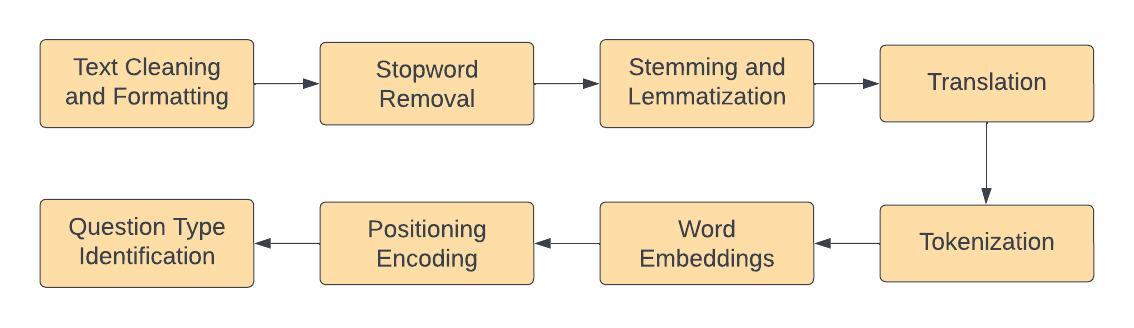} 
    \caption{Data Preprocessing Process}
    \label{fig:Preprocessing}
\end{figure}

\setlength{\parindent}{1em}

Text Cleaning and Formatting: All the papers prioritize cleaning and formatting the textual data to ensure consistency and remove any unwanted noise. This typically involves eliminating extraneous symbols, punctuation marks, and non-alphanumeric characters that could disrupt subsequent processing steps. This initial cleaning step sets the stage for uniform and standardized text data.

\begin{figure}[htbp] 
    \centering
    \includegraphics[width=150px]{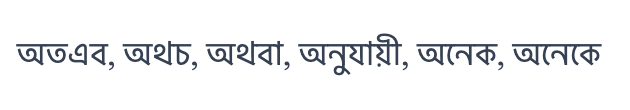} 
    \caption{Stop Words Example}
    \label{fig:StopWords}
\end{figure}

Stopword Removal: Many of the papers acknowledge the importance of removing stopwords—common words. These stopwords are often removed as they don't carry substantial semantic meaning and can introduce noise into the data. Their removal helps streamline text analysis and reduce dimensionality. The sample of stop-words are shown in Fig.\ref{fig:StopWords}.

Stemming and Lemmatization: Stemming or lemmatization techniques are used in some papers to reduce words to their base or root forms. This process helps in consolidating variations of words and reduces the diversity of vocabulary, making it easier for models to recognize and match similar words.

Translation: Due to the scarcity of Bengali-language datasets, some papers resort to translating existing datasets from English to Bengali. This translation process requires careful handling to ensure the translated questions and answers retain their intended meanings and contexts. Proper translation is crucial to maintain data quality.

Tokenization: Tokenization is a key preprocessing step where text is segmented into individual tokens, typically words or subwords. Tokenization is essential for feeding text data into neural network models, as it converts raw text into a format suitable for analysis.

Word Embeddings: Several papers leverage word embeddings, such as Word2Vec, to represent words as vectors. Word embeddings capture semantic relationships between words based on their co-occurrence in a large corpus. This enhances the models' understanding of word meanings and helps improve the overall performance of the question-answering systems.

Positioning Encoding: In one of the papers, positioning encoding is introduced. This technique assigns unique positions to words in a sentence, enhancing the model's comprehension of word order. This is particularly crucial for capturing contextual information, as it ensures that the model recognizes the sequential structure of the text.

Question Type Identification: Some datasets used in these papers include question types. Identifying and categorizing these question types is part of the preprocessing, as it enables the analysis of system performance across different question categories.
\setlength{\parindent}{0em}

These data preprocessing techniques collectively prepare the text data for subsequent stages, such as training neural network models. They help ensure that the data is clean, standardized, and semantically meaningful, facilitating accurate and effective question-answering system development in the context of the Bengali language.


\subsection{Methodology and Models} The research papers collectively employ a range of methodologies and advanced models to address the challenge of building effective question-answering (QA) systems in Bengali. These methodologies focus on leveraging deep learning techniques, transfer learning, and innovative approaches. Pre-trained models like multilingual BERT \cite{mBERT}, DistilBERT \cite{DistilBERT}, and RoBERTa \cite{roberta} play a pivotal role. The studies explore both zero-shot transfer learning and fine-tuning strategies. The utilization of Positioning Encoding and Word2Vec \cite{word2Vec} word representation techniques enhances word comprehension and semantic relationships within models.  Deep neural network architectures such as LSTM \cite{lstm}, Bi-LSTM \cite{bi_lstm}, RNN \cite{rnn}, ELECTRA \cite{electra}, and BERT \cite{bert},  has been utilized. Synthetic datasets, created through translation and fuzzy answer matching, provide the basis for model training. Additionally, a comprehensive set of experiments compares the performance of various models, revealing the most effective strategies for adapting pre-trained models to the Bengali QA context. The findings underscore the potential of transfer learning and model fine-tuning in bridging the gap between the availability of annotated data and the development of accurate QA systems for Bengali.

\subsection{Evaluation Metrics and Performance Analysis} The research papers collectively employ comprehensive evaluation metrics to gauge the effectiveness and efficiency of the proposed question-answering (QA) systems. Metrics such as Mean Reciprocal Rank (MRR), precision, recall, and F1 score are utilized to quantitatively assess the models' performance. Through systematic performance analysis, the studies offer insights into the models' ability to accurately retrieve answers and match user queries, shedding light on the systems' real-world viability.

\subsection{Results and Analysis:}
The results of the diverse models employed in these studies have yielded insightful findings, providing a comprehensive view of the state of Bengali question-answering systems. In the comparison between Bangla and English-based question-answer systems, a Sequence to Sequence LSTM-based approach with attention mechanisms was proposed. After training over 35 epochs, the English model exhibited superior performance with an accuracy of 90.16\%, outperforming the Bangla model's accuracy of 87.36\%. While this performance gap highlights the current disparity, it also underscores the need for further research and development to enhance Bengali question-answering systems. Similarly, in the study on Bengali Factoid Question Answering (BFQA), a pipeline architecture was introduced, achieving substantial inter-rater agreement with kappa scores of 0.91, 0.85, and 0.89 for question class, expected answer type, and question topical target identification, respectively. The automatic context-based QA system employed Sequence-to-Sequence LSTM with an attention mechanism, significantly lowering training loss to 0.003. Moreover, this system achieved an average precision of 0.35 and recall of 0.65, resulting in an F score of 0.45, demonstrating its effectiveness in information retrieval and understanding. In the context of sentence similarity, utilizing universal sentence encoders, the Pearson correlation coefficient was calculated at 0.41, highlighting the model's ability to gauge text relatedness. This finding is crucial for various natural language understanding tasks. Notably, the paper also focused on constructing an End-To-End Neural Network for Paraphrased Question Answering in Bengali. The best accuracy attained using LSTM with Word2Vec representation was 100\%. However, the standout result came from the use of DistilBERT, which exhibited the highest accuracy of 98.86\% among the various models, showcasing the potential of transfer learning for Bengali QA tasks and emphasizing the importance of leveraging pre-trained models. The results and summaries of papers are shown in Table \ref{tab:table1}.

\begin{table}[htbp]
\centering
\caption{Summary of Papers on Bengali Question Answering}
\begin{tabular}{|p{2.4cm}|p{1.7cm}|p{2.9cm}|}
\hline
\textbf{Year and Title} & \textbf{Models Used} & \textbf{Results} \\
\hline
[2021] SONDHAN: A Comparative Study of Two Proficiency... & Seq2Seq, LSTM, Attention Mechanism & 
\begin{itemize}
    \item English Model: Prediction Accuracy: 90.16\%
    \item Bangla Model: Prediction Accuracy: 87.36\%
\end{itemize} \\
\hline

[2014] BFQA: A Bengali Factoid Question Answering System... & LSTM, BiLSTM & 
\begin{itemize}
    \item Mean Reciprocal Rank (MRR) value: 0.32
    \item Factoid QA system for Bengali
\end{itemize} \\
\hline

[2020] An Approach for Bengali Automatic Question Answering... & Seq2Seq, BiLSTM, Attention Mechanism & 
\begin{itemize}
    \item Reduced training loss to 0.003
    \item Close-domain approach in Bengali
\end{itemize} \\
\hline

[2019] Design and Development of Question Answering System... & Various NLP techniques & 
\begin{itemize}
    \item Precision: 0.35, Recall: 0.65, F Score: 0.45
    \item Textual Entailment Module (TE) for validation
\end{itemize} \\
\hline

[2021] Bengali Context–Question Similarity Using Universal... & Universal Sentence Encoder & 
\begin{itemize}
    \item Pearson Correlation value: 0.41
    \item Context-query similarity with cosine distance
\end{itemize} \\
\hline

[2020] End-To-End Neural Network for Paraphrased Question... & LSTM, GRU, Word2Vec & 
\begin{itemize}
    \item Achieved accuracy: LSTM + Word2Vec (100\%)
    \item Zero-shot and fine-tuned approaches
    
\end{itemize} \\
\hline
[2020] Deep Learning-Based Question Answering System in Bengali... & BERT, DistilBERT, RoBERTa & 
\begin{itemize}
    \item EM Accuracy score: 52.5
    \item F1 Accuracy score: 69.21
\end{itemize} \\
\hline

[2021] Bangla Intelligence Question Answering System... & TF-IDF model, Singular Value Decomposition (SVD) & 
\begin{itemize}
    \item Cosine Similarity accuracy: 93.22
    \item Jaccard Similarity accuracy: 82.64
    \item Naive Bayes classification accuracy: 91.34
  
\end{itemize} \\
\hline

[2022] BanglaRQA: A Benchmark Dataset for Under-resourced Bangla... & BanglaT5, mT5 ,BanglaBERT, mBERT & 
\begin{itemize}
    \item Exact Match score: 62.42%
    \item F1 score: 78.11%
  
\end{itemize} \\
\hline

[2022] Reading comprehension based question answering system...
 & LSTM, Bi-LSTM, RNN, ELECTRA, BERT & 
\begin{itemize}
    \item BERT accuracy: 87.78\% 
    \item ELECTRA accuracy: 86.5\%.

\end{itemize} \\
\hline
\end{tabular}
\label{tab:table1}
\end{table}

\subsection{Limitations:} The papers discussing the development of Bengali question-answering systems collectively grapple with several limitations. Challenges arise from the scarcity of high-quality, large-scale Bengali datasets, often leading to the translation of English or other language datasets, which can introduce errors. Resource constraints, such as limited computational power and human annotators, hinder the scope and complexity of experiments. English models consistently outperform Bengali ones, reflecting a performance gap that underscores the need for further research. Bengali's low-resource status in the realm of natural language processing presents additional hurdles. The narrow focus on specific question types or domains may not adequately cover real-world use cases with varied context. Moreover, discussions about scalability, generalization, and broader applications remain somewhat limited in these papers. Nevertheless, these studies offer valuable insights and methodologies for advancing Bengali question-answering systems while acknowledging the ongoing challenges in this field.

\section{Conclusion}
In conclusion, the realm of Bangla Question Answering Systems using Natural Language Processing (NLP) has seen remarkable progress through a series of insightful studies. Each of the seven papers discussed above has contributed significantly to addressing the challenges associated with this specialized domain. Nusrat Nabi et al.'s research offers a deep understanding of Sequence to Sequence LSTM-based models, while Somnath Banerjee et al.'s work tackles the scarcity of Bengali QA corpora, leading to innovative solutions. Md. Rafiuzzaman Bhuiyan et al. emphasizes the power of sequence-to-sequence architecture \cite{Sequence_to_Sequence}, while Samina Tasnia Islam et al.'s approach to answer retrieval is both innovative and informative. Furthermore, M. Keya et al.'s exploration of semantic similarity and sentence matching enriches the field, and the paper by Bangla and English-based question-answer systems offers valuable insights into comparative analysis. Lastly, the deep learning techniques employed by the "End-To-End Neural Network for Paraphrased Question Answering Architecture with Single Supporting Line in Bangla Language" paper serve as a testament to the potential of transfer learning. Together, these studies signify a significant step forward in the development of Bangla NLP applications. The dedication to addressing the unique challenges posed by the Bengali language, coupled with the integration of cutting-edge methodologies, demonstrates a commitment to bridging the language-technology gap. As these advancements continue to unfold, they hold the promise of enhancing the accessibility, usability, and effectiveness of Bangla Question Answering Systems. The insights garnered from these studies lay a solid foundation for future researchers and developers to build upon, ultimately contributing to the evolution of technology that resonates deeply with the linguistic and cultural nuances of the Bengali language.

\bibliographystyle{ieeetr}
\bibliography{refs.bib}
\end{document}